\documentclass[lettersize,journal]{IEEEtran}
\usepackage{amsmath,amsfonts}
\usepackage{algorithmic}
\usepackage{algorithm}
\usepackage{array}
\usepackage[caption=false,font=normalsize,labelfont=sf,textfont=sf]{subfig}
\usepackage{textcomp}
\usepackage{stfloats}
\usepackage{url}
\usepackage{verbatim}
\usepackage{graphicx}
\usepackage{cite}

\usepackage{amsmath}
\usepackage{booktabs}

\usepackage{amsthm,amsmath,amssymb}
\usepackage{mathrsfs}

\usepackage{multirow}

\usepackage{caption}

\begin{document}

\title{Prompt-Learning for Short Text Classification}

\author{Yi Zhu, Xinke Zhou, Jipeng Qiang, Yun Li, Yunhao Yuan, and Xindong Wu, ~\IEEEmembership{Fellow,~IEEE}
        % <-this % stops a space
\thanks{This paper was produced by the IEEE Publication Technology Group. They are in Piscataway, NJ.}% <-this % stops a space
\thanks{Manuscript received April 19, 2021; revised August 16, 2021.}}

\markboth{Journal of \LaTeX\ Class Files,~Vol.~14, No.~8, August~2021}%
{Shell \MakeLowercase{\textit{Zhu et al.}}: Prompt-Learning for Short Text Classification}


\maketitle

\begin{abstract}

In the short text, the extremely short length, feature sparsity, and high ambiguity pose huge challenges to classification tasks. Recently, as an effective method for tuning Pre-trained Language Models for specific downstream tasks, prompt-learning has attracted a vast amount of attention and research. The main intuition behind the prompt-learning is to insert the template into the input and convert the text classification tasks into equivalent cloze-style tasks. However, most prompt-learning methods expand label words manually or only consider the class name for knowledge incorporating in cloze-style prediction, which will inevitably incur omissions and bias in short text classification tasks. In this paper, we propose a simple short text classification approach that makes use of prompt-learning based on knowledgeable expansion. Taking the special characteristics of short text into consideration, the method can consider both the short text itself and class name during expanding label words space. Specifically, the top $N$ concepts related to the entity in the short text are retrieved from the open Knowledge Graph like Probase, and we further refine the expanded label words by the distance calculation between selected concepts and class labels. Experimental results show that our approach obtains obvious improvement compared with other fine-tuning, prompt-learning, and knowledgeable prompt-tuning methods, outperforming the state-of-the-art by up to 6 Accuracy points on three well-known datasets.

\end{abstract}


\section{Introduction}

\IEEEPARstart{W}{ith} the rapid development of web service, short-texts, which are posted at unprecedented rates, accentuate both the importance of learning tasks and the challenges posed by the inherent properties such as the extremely short length, feature sparsity, and high ambiguity. In recent decades, short text classification has attracted a vast amount attention and research from multiple disciplines \cite{sun2012short,cui2021pre}, the advances in short text data processing have far-ranging consequences in practical applications like Twitter \cite{sriram2010short}, Facebook \cite{faqeeh2014cross}, and Microblog \cite{liu2010short}.

Existing methods for short text classification can roughly be categorized into two classes: sole source and external knowledge base methods. The methods based on sole source extend the feature space by the rules or statistical information hidden in the current short texts \cite{kim2014convolutional,lai2015recurrent}. The sole source methods still face the severe data sparsity problem, the methods based on external knowledge are applied widely in recent years, which extends the feature space by external open knowledge base \cite{chen2019deep,xu2019incorporating}. However, most existing methods based on external knowledge rely on large-scale training instance data to formalize the model, which leads to high costs in collecting eligible training data and performance degradation in few-shot learning.

Recently, Pre-trained Language Models (PLMs) have attained much attention and remarkable improvements in a series of downstream Natural Language Processing (NLP) tasks, such as text classification \cite{minaee2021deep}, question answering \cite{yang2019data}, machine translation \cite{weng2020acquiring}, and lexical simplification \cite{qiang2020lexical}. PLMs can learn syntactic \cite{goldberg2019assessing}, semantic \cite{ma2019universal} and structural \cite{jawahar2019does} information about language. To adapt the versatile knowledge contained in PLMs to various NLP tasks, the fine-tuning method with extra classifier has been applied widely to stimulate and exploit rich knowledge in PLMs and has achieved excellent performance in downstream tasks \cite{han2021pre}.

Among all the fine-tuning approaches, inspired by GPT-3 that uses the information provided by the prompts in few-shot learning and achieves substantive results \cite{brown2020language}, prompt-learning fill the input statements into the natural language template and adapt the masked model, which formalized regarded the downstream NLP task as cloze-style tasks. For example, to classify the topic of sentence $x$ as "Call them the 'Nightmare Team'." into the "Sports" category, the template can be noted as "$x$, a [MASK] question", and prompt-learning predict the probability that the word "sports" is filled in the "[MASK]". Compared with the previous fine-tuning approaches, no additional neural layer is needed in prompt-learning, and excellent performance has been achieved even in the scenario of few-shot or zero-shot learning. In the prompt-learning, the mapping from label words (e.g. sports, association, basketball et al.) to the category (e.g. SPORTS) can effectively alleviate the discrepancy between text and label space, which is called the automatic selection of label words \cite{gao2020making} or the verbalizer \cite{schick2020automatically}. This strategy of constructing label word mapping has been proved to be effective in achieving more desirable text classification performance \cite{schick2020exploiting}.

However, most prompt-learning methods either expand label words manually \cite{schick2020s} or only consider the class name for knowledge incorporating in cloze-style prediction \cite{schick2020automatically}. The manually designed label words are obviously limited with the prior knowledge, which may induce omissions and bias for knowledge expansion. Some other works try to incorporate external knowledge and denoise expanded label words for text classification \cite{hu2021knowledgeable}. However, since the special characteristics of short text, such as the extremely short length, feature sparsity, and high ambiguity, are totally different from conventional text, such a knowledgeable method only consider the class name and ignore the entity and concept information in the short text, which achieved unsatisfying results in short text classification.

Therefore, we present an intuitive and innovative idea for short text classification in this paper. We exploit recent advances in the prompt-learning model \cite{ding2021openprompt} based on knowledgeable expansion in the few-shot scenario. Taking the special characteristics of short text into consideration, the proposed Prompt-Learning approach for Short Text classification (PLST) incorporated both the short text itself and external knowledge from open Knowledge Graph like Probase to extend label words space. More specifically, the top $N$ concepts concerning the entities in the short text are firstly retrieved from open Knowledge Graph such as Probase. Then the distance is calculated between retrieved concepts and class labels in embedding space for label words refinement. The advantage of our method is that it generates more effective label words by considering the short text itself, not just the class name.

Here, we give an example shown in Figure 1 to illustrate the advantage of our method PLST. For one sentence "Ford cuts production while Chrysler's sales rise." and template "This topic is about [mask]", the label word space in Prompt-tuning \cite{ding2021openprompt} only includes the class name 'business', which refers to that only predicting the word "business" for the [MASK] token is regarded as correct regardless of other relevant words. The expanded label words generated by Knowledgeable Prompt-tuning \cite{hu2021knowledgeable} contain plenty of words as {commerce, trade, market, antique, purchase,...}, which only related to the class name 'business' without paying attention to the original sentence. The expanded label words generated by our PLST are not related to the class name, but also can expand knowledge from the original sentence. Then, by considering the distance as a filter for label words refinement, 'company' and 'manufacturer' are selected as the expansion for the class business. In this case, the expanded label words space {business, company, manufacturer,...} is more accurate and more reliable than other label words space.

\begin{figure}[htbp]
\centerline{\includegraphics[width=3in,height=2.59in]{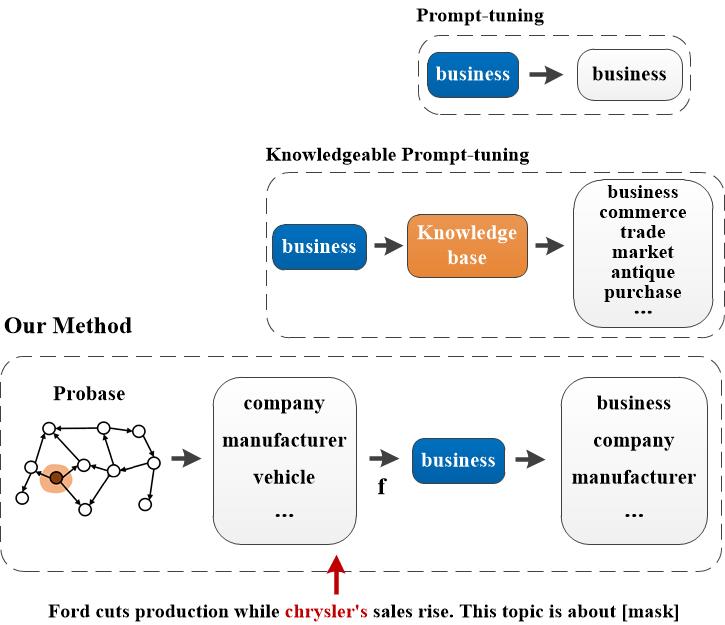}}
\caption{Comparison of label words expansion with different approaches. Given one sentence "Ford cuts production while Chrysler's sales rise." and template "This topic is about [mask]", the expanded label words are generated by our method PLST and the state-of-the-art two baselines based Prompt-tuning \protect\cite{ding2021openprompt} and Knowledgeable Prompt-tuning \protect\cite{hu2021knowledgeable}.}
\label{fig}
\end{figure}

The contributions of our paper are summarized as follows:

(1) Our PLST is a novel prompt-learning-based method for short text classification, which can take full advantage of prompt-learning to bridge the objective form gap between pre-training and fine-tuning. Compared with existing methods, our PLST can achieve more desirable performance, since it considers both the short text and class name during expanding label words space.

(2) Our PLST is a simple and effective short text classification method. 1) Simple: many steps used in existing short text classification methods have been eliminated from our method, e.g., co-occurrence words searching and the entire model tuning. 2) Effective: it obtains new state-of-the-art results on three benchmarks.

(3) To our best knowledge, this is the first attempt to apply prompt-learning models on short text classification tasks. The code to reproduce our results is available at https://github.com/anonymous.
%https://github.com/z8d1177/PLST.

\section{Related Work}

The last decades have witnessed a vast amount of interest and research on short text, which has played a crucial role in many real-world scenario applications. Recently, PLMs and prompt-learning have attained substantial performance in a series of downstream NLP tasks, and a prompt-learning strategy that incorporated knowledge for verbalizer is proposed in this paper. In this section, we review the literature pertaining to short text classification, prompt-learning, and verbalizer construction respectively.

\subsection{Short text classification}

Short text classification has provoked a vast amount of attention and research in recent decades \cite{chen2019deep}, which has played an import role in many practical applications like sentiment analysis \cite{song2020sacpc}, dialogue systems \cite{lu2013deep}, and user intent understanding \cite{hu2009understanding}. Short text classification aims to process texts with very short length, usually no more than 100 characters, such as blog content \cite{liu2010short}, online comments \cite{chen2020verbal}, news title \cite{wu2020mind} and so on.

Existing short text classification methods can be roughly divided into two categories: sole source methods and external knowledge base methods. The methods based on sole source expand the feature space by disentangling explanatory factors of variations behind the current short text \cite{kim2014convolutional}. For example, Lai et al. proposed a recurrent convolutional neural network to capture contextual information and the key components for text classification \cite{lai2015recurrent}. Bollegala et al. proposed a ClassiNet network to predict missing features for addressing feature sparseness problems, unlabeled data are utilized to generalize word co-occurrence graphs, and the relations between features and short text are explored \cite{bollegala2018classinet}. Hao et al. proposed a Mutual-Attention Convolutional Neural Networks for Chinese short text classification, which integrates word and character-level features without losing feature information \cite{hao2020chinese}.

However, the sole source methods still face the severe data sparsity problem, the methods based on external knowledge are applied widely in recent years, which extends the feature space by the external open knowledge base. For example, Chen et al. proposed to retrieve knowledge from external knowledge source to enhance the semantic representation of short texts, and attention mechanisms are introduced in this method to acquire the weight of concepts \cite{chen2019deep}. Xu et al. proposed a hybrid model to incorporate context-relevant knowledge into a convolutional neural network for short text classification \cite{xu2019incorporating}. Yang et al. proposed a heterogeneous information network to incorporate additional information and their relations from open Knowledge Base, which can address the semantic sparsity problem in short text classification \cite{yang2021hgat}. Although these methods can obtain fairly good results in short text classification, they all rely on the large-scale training instance data to formalize the model, which leads to high costs in collecting eligible training data and performance degradation in few-shot learning.

\subsection{Prompt-learning}

Recently, the fine-tuned Pre-trained Language Models have achieved tremendous success in various NLP tasks, such as question answering \cite{yang2019data,adiwardana2020towards}, text classification \cite{minaee2021deep,ding2021prototypical}, machine translation \cite{zhu2020incorporating,weng2020acquiring}, and lexical simplification \cite{bao2020enhancing,qiang2020lexical}. PLMs can learn syntactic \cite{goldberg2019assessing}, semantic \cite{ma2019universal} and structural \cite{jawahar2019does} information about language. For example, Ye et al. constructed a graph convolutional network for short text, the word and document nodes trained by the GCN and vector generated by the BERT are input into the BiLSTM classifier for short text classification \cite{ye2020document}. Sun et al. conducted extensive experiments to investigate the different approaches to fine-tuning BERT and achieved state-of-the-art performance on short text classification tasks \cite{sun2019fine}.

However, the huge gap between pre-training and fine-tuning still prevents downstream tasks from fully utilizing pre-training knowledge. To this end, inspired by GPT-3 \cite{brown2020language}, prompt-learning has been proposed to transfer downstream tasks as some cloze-style objectives and achieved superior performance, especially in few-shot learning \cite{liu2021pre}. Along this line, many hand-crafted prompts have been made in various tasks, such as knowledge probing \cite{petroni2019language,davison2019commonsense}, relation extraction \cite{han2021ptr}, entity typing \cite{ding2021prompt} and text classification \cite{wang2019superglue,hu2021knowledgeable}. For example, Han et al. applied logic rules to construct prompts with several sub-prompts on relation classification, which consistently outperforms existing state-of-the-art baselines without introducing any additional model layers, manual annotations, and augmented data \cite{han2021ptr}. Ding et al. proposed a prompt-learning method on fine-grained entity typing, the entity types are extracted without over-fitting by performing distribution level optimization \cite{ding2021prompt}. Furthermore, to avoid time-consuming and labor-intensive prompt design, a series of automatic prompt generation methods have been explored recently \cite{li2021prefix,lester2021power}. For example, Shin et al. proposed an automatically generated prompts method to elicit knowledge from language models on sentiment analysis and textual entailment, which outperforms manual prompts with less human effort \cite{shin2020autoprompt}. Hambardzumyan et al. proposed an automatic prompt generation method to transfer knowledge from large PLMs to downstream tasks by appending embeddings to the input text, which significantly outperforms baselines in a few-shot setting \cite{hambardzumyan2021warp}.

\subsection{Verbalizer construction}

In the prompt-learning, the verbalizer refers to a projection from label words  (e.g. sports, basketball et al.) to the category (e.g. SPORTS), which has been proven to be an important and effective strategy for alleviating the discrepancy between text and label space \cite{schick2020automatically}. The hand-crafted verbalizers have achieved sound performance in text classification and other NLP tasks. For example, Schick et al. proposed to use pairs of cloze question patterns and manually designed verbalizers for leveraging the knowledge contained within PLMs for downstream tasks \cite{schick2020exploiting}. However, the manually designed verbalizers are highly impacted by the prior knowledge and incur omissions and bias for knowledge expansion.

Since the hand-crafted verbalizers require enough training and validation sets for optimization, a series of the automatic verbalizer construction methods in the prompt-learning are presented \cite{gao2020making,schick2020automatically}. For example, Wei et al. proposed a prototypical network to generate prototypical embeddings for different labels in the feature space, which summarized the semantic information of labels to form a prototypical prompt verbalizer \cite{wei2022eliciting}. However, synonyms of category names are more likely to be expanded instead of diverse and comprehensive label words in these methods. To denoise expand label words in automatic verbalizer, some other works try to select related words from external knowledge base \cite{hu2021knowledgeable}. Such a method can greatly enhance the semantics of labels, but due to a large number of useless words being extracted during the stage of verbalizer construction, the verbalizer refinement is difficult and leads to unsatisfied results in few-shot learning. Compared to the previous methods, in this paper, we proposed a prompt-learning strategy to significantly improve the performance of short text classification. Not only class name but also the short text itself are taken into consideration for knowledgeable expansion.

\section{Methodology}

In this section, we briefly summarize the prompt-learning method, and then label words set construction, cluster, and refinement, and text classification are described successively.

\begin{figure*}[htbp]
\centerline{\includegraphics[width=5in,height=3.57in]{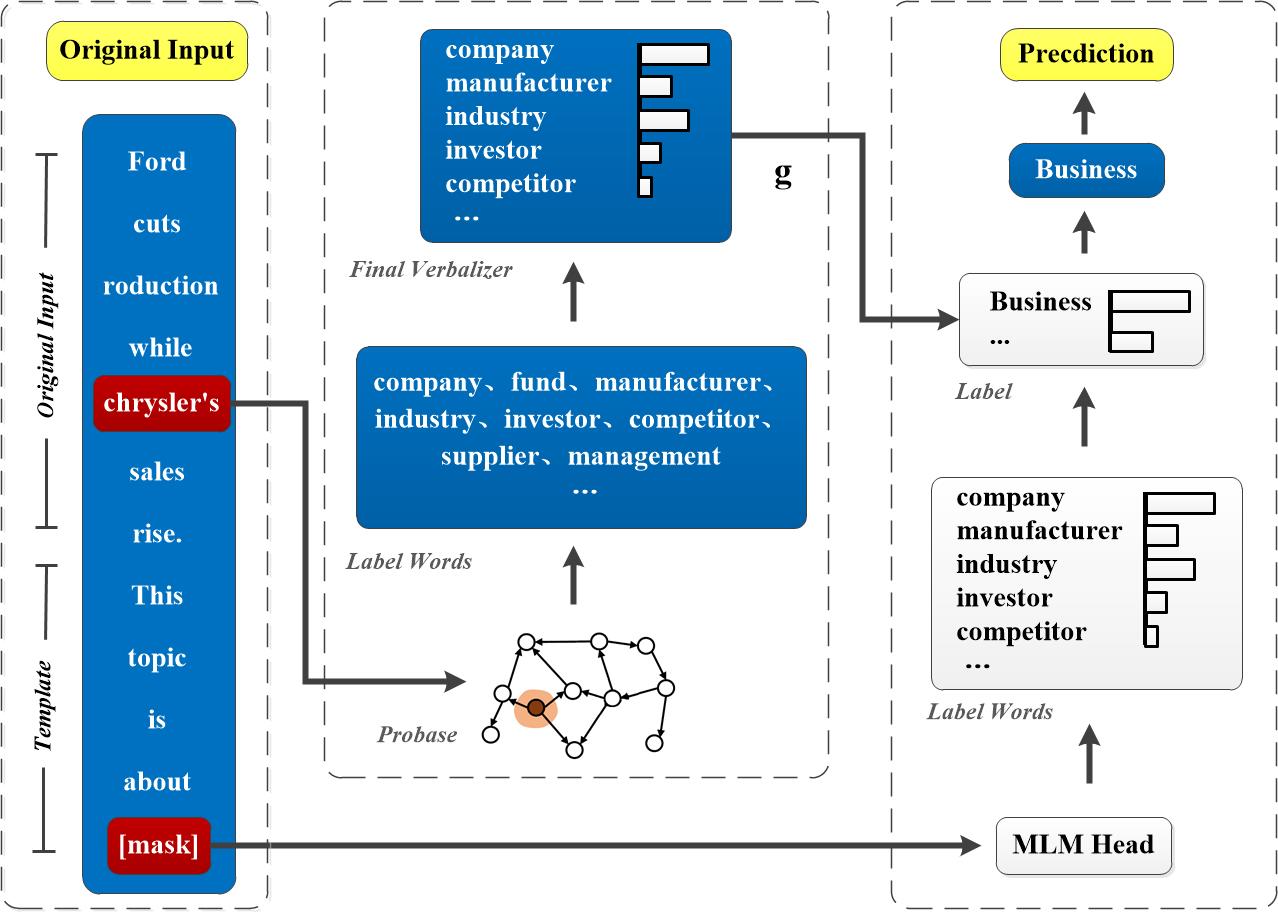}}
\caption{The illustration of our PLTS for verbalizer construction. MLM is short for Masked Language Modeling. The left part refers to the input text: "Ford cuts production while Chrysler's sales rise." with the template: "This topic is about [mask]". The center and the right part are the label words set construction by the short text itself and class name respectively. The red part in Probase refers to the concept node corresponding to 'Chrysler' in the input text.}
\label{fig}
\end{figure*}

\subsection{The prompt-learning method}

In the prompt-learning method, the input statements are formalized as the natural language template, and the text classification tasks are regarded as the close-style tasks. For example, in topic classification, assuming we need to classify the sentence $x$: "yukos cuts output to save money" into the label $y_1 = BUSINESS$ or $y_2 = SPORTS$, the template can be noted as:

$$
x_p  = \left[ {{\rm{CLS}}} \right]x{\rm{, a}}\left[ {{\rm{MASK}}} \right]{\rm{question}}
$$

Given the input $x = \left\{ {x_1 ,...,x_n } \right\}$, which is classified into a category with label $y \in Y$, the label word set is denoted as $V_y  = \left\{ {v_1 ,...,v_n } \right\}$, where $V_y$ is a subset of the whole vocabulary $V$, i.e., $V_y \in V$, and $V_y$ is mapped into a category with label $y$. In PLMs $M$, the probability that each word $v$ in $V_y$ is filled in the $\left[ {{\rm{MASK}}} \right]$ can be shown as $ p\left( {\left[ {{\rm{MASK}}} \right] = v \in V_y |x_p } \right) $. Thus, the text classification tasks can be transferred into a probability calculation problem of label words, which can be computed as \eqref{proCal}:

\begin{equation}
p\left( {y \in Y|x} \right) = p\left( {\left[ {{\rm{MASK}}} \right] = v \in V_y |x_p } \right)
\label{proCal}  \end{equation}

In the above mentioned example, if the calculated probability of $V_1  = \left\{ {business} \right\}$ for $y_1 = BUSINESS$ is larger than $V_2  = \left\{ {sports} \right\}$ for $y_1 = SPORTS$, it indicates that the sentence $x$ is classified into the category $BUSINESS$.

In the scenario of automatic expansion of label words or verbalizer construction, the $V_y$ related to a special category with label $y$ is expanded, such as $V_1  = \left\{ {business} \right\}$ can be expand as $V_1  = \left\{ {business, commerce, company,...} \right\}$ in the above example, which can obviously improve the short text classification performance with prompt-learning. In this paper, taking the special characteristics of short text into consideration, we consider both the short text itself and the class name for verbalizer construction. The top $N$ concepts related to the entity in the short text are retrieved from the open Knowledge Graph like Probase, and we further refine the expanded label words by the distance calculation between selected concepts and class labels. The details are in the following sections.

\subsection{Label Words Set Construction}

In short text classification, the crucial problem for label words expansion is the hierarchical label space, which refers to multiple aspects and granularities. For example, "country", "province", and "city" are multi-level and related words, and they may be all fit the predicting masked words in the template of prompt-learning. To this end, we select Probase\footnote{https://concept.research.microsoft.com/.} as the external knowledge source, which is an open Knowledge Graph constructed by Microsoft. Probase specifies the probability of each instance belonging to the concept, and the concepts with relatively small correlations will be abandoned in the process of label words expansion, which reduces the difficulty in verbalizer refinement. The label name $y$ of each topic is utilized as anchor words, and the top $N$ concepts are retrieved from Probase ranked with probabilities, which can be denoted as $N\left( v \right)$. Thus, the expanded label words set can be represented as $V_y  = \{ y\}  \cup N\left( v \right)$, which is a verbalizer mapping label words set to a special category, some expanded label words in our PLST are listed in Table 1.

\newcommand{\tabincell}[2]{\begin{tabular}{@{}#1@{}}#2\end{tabular}}  %表格自动换行
\begin{table*}
\centering
\caption{Examples of the expanded label words}
\begin{tabular}{ccc}
\toprule
\textbf{Dataset} & \textbf{Label} & \textbf{Label Words Set}\\
\hline
\multirow{2}*{\tabincell{c}{AG's News}} & $BUSINESS$ & industry, company, manufacturer, investor, provider, giant, fund, insurance, ...\\

~ & $SPORTS$ & rivalry, playing, fifa, games, winners, athlete, career, soccer, ... \\
\hline

\multirow{4}*{\tabincell{c}{Snippets}} & $POLITICS$ & religious, cultural, province, minority, terrorist, nationality, politician, government, ...\\

~ & $HEALTH$ & nutrition, care, prevention, cancer, patient, fitness, healthcare, disease, hiv, ...\\

~ & $COMPUTERS$ & server, file, editing, software, format, Microsoft, code, advanced, tool, ... \\

~ & $SPORTS$ & club, nba, soccer, basketball, players, football, teams, baseball, standings, match, ... \\

\bottomrule
\end{tabular}
\label{tab:booktabs}
\end{table*}

\subsection{Cluster and Refinement}

Although the expanded label words set is filtered by concepts probabilities firstly, there are still many useless and noise words since the gap between the pre-training model and concepts in the knowledge graph. Therefore, it is necessary to further refine the label words set for retaining high-quality and removing low-relevance words.

In the scenario of few-shot learning, the refinement should consider the possible impact of each expanded label word on the classification. Since the verbalizer is a projection between label words set and the category label space, in the embedded space, the distance $dist\left( {V_y ,y} \right)$ between each expanded label words set and each label name $y$ is calculated. In this way, the higher-probability words for classification are clustered into the label name of each topic, which not only considers the hierarchical concept itself but also can fit the label word of the special category. Finally, we select the top $M$ words from $dist\left( {V_y ,y} \right)$ for each category as verbalizer refinement, excluding the morphological derivations of $y$.

\subsection{Text Classification}

After the final verbalizer construction, we need to map the predicted probability on each refined label word to a special class, which can be noted as an objective function $g$ for verbalizer utilization. Due to that each word in the final verbalizer can be assumed to contribute equally for predicting, the average of the predicted scores is used for text classification, i.e., $g$ can be calculated as \label{predict}:

\begin{equation}
\arg \max _{y \in Y} \frac{1}{{\left| {V_y } \right|}}\sum\limits_{v \in V_y } {p\left( {\left[ {{\rm{MASK}}} \right] = v|x_p } \right)}
\label{predict}  \end{equation}

Suppose that there is a sentence "Ford cuts production while Chrysler's sales rise." and the corresponding topic is $BUSINESS$, we can obtain the label words in verbalizer as \{company, manufacturer, industry, investor, competitor, ...\}. The whole framework of our proposed PLST is illustrated in Figure 2, and we can see the expanded label words not only have a strong correlation with the category $BUSINESS$ but also hold hierarchical and multi-granularities properties of the topic. If we adopt the existing state-of-the-art method based on Knowledgeable Prompt-tuning proposed by \cite{hu2021knowledgeable}, the expanded label words are \{commerce, trade, market, antique, purchase,...\}. Very obviously, our method generates a better label words set.

\section{Experiments}

In this section, we conduct experiments on three datasets to evaluate the effectiveness of our proposed method in short text classification.

\subsection{Datasets and Templates}

We conduct experiments on three short text datasets: AG's News \cite{zhang2015character}, Snippets \cite{phan2008learning} and News Title \cite{linmei2019heterogeneous}. Notably, we only pick the news title of AG's News as short text in the experiments. The statistics of each dataset are listed in Table 2.

\begin{table}[htbp]
\begin{center}
\caption{The details of all three datasets}
\begin{tabular}{c|c|c|c}
\hline
Dataset & Class & Test Size & Avg.Len\\
\hline
AG news & 4 & 7600 & 7\\
\hline
Snippets & 8 & 2280 & 18\\
\hline
News title & 7 & 6521 & 8\\
\hline
\end{tabular}
\end{center}
\end{table}

In the experiments, the manual templates are used due to the topic templates are simple and effective. To test the influence of different topic templates, further experiments are conducted on all three datasets in Section 4.5.

\subsection{Compared Methods and Experiment Settings}

The BERT model, fine-tuning, regular prompt-learning, and Knowledgeable Prompt-tuning methods are conducted to demonstrate the effectiveness of our proposed method.
%Moreover, the state-of-the-art short text classification methods are also included since they have achieved sound performance.

\begin{itemize}
\item BERT + Fine-tuning (BERT + FT) \cite{devlin2018bert}. This method first obtains the hidden embedding of [CLS] token through Bert, and then inputs this hidden embedding into the classification layer to predict.
\item Regular prompt-learning (PL) \cite{liu2021pre}. The regular prompt-learning fills the input statements into a hand-crafted template, and only the category name is used to form the label word space. For fairness, the hand-crafted templates in PL are consistent with our method.
\item Knowledgeable Prompt-tuning method (KPT) \cite{hu2021knowledgeable}. KPT method expands the verbalizer in prompt-tuning with the external Knowledge Base. The verbalizer construction, refinement and utilization are used to incorporate external knowledge for prediction.
\end{itemize}

\textbf{Experiment Settings:} In our experiments, we conduct 5, 10, 20-shot as $k$-shot learning experiments. Specifically, $k$ and 1000 instances of each category are sampled from the original training set to form a new training set and support set respectively. It is worth mentioning that all experimental results are obtained by repeating five experiments and taking the average value. Moreover, accuracy is adopted in all the experiments as the test metric. For the PLMs, we use ${\rm{RoBERTa}}_{{\rm{LARGE}}}$ \cite{liu2019roberta} for all experiments. The OpenPrompt \cite{ding2021openprompt} is adopted for the implementation of PL and KPT.

\subsection{Experimental Results}

All the experimental results on three datasets are recorded in Table 3. The following insightful observations can be listed from experimental results:

(1) Generally, as the experiments vary from 5-shot to 20-shot, the performance of all fine-tuning methods has improved, which reveals the increase of labeled instances number can improve the results in few-shot learning.

(2) Our method and KPT achieve more desirable performance than other fine-tuning methods in most cases, which indicates the effectiveness of expanding the verbalizer in prompt-tuning with the external Knowledge Base.

(3) The results of the KPT are not stable especially in the News Title dataset, which shows that KPT cannot incorporate appropriate knowledge in some datasets. Moreover, we can observe that the performance of KPT is even worse than regular prompt-learning on the News Title dataset, it further demonstrates that the expanded label words by KPT may have some bias and do not have enough coverage in short texts.

(4) Overall, our PLST performs best in most cases, which validates the effectiveness of incorporating external knowledge for extending label words space to address the inherent problems of short text classification. It should be noted that our method can achieve stable performance in all three datasets, which shows the effectiveness of knowledgeable expansion with concepts retrieval and refinement.

\begin{table}
\centering
\caption{The results of accuracy on three datasets.}
\begin{tabular}{ccccc}
\toprule
\textbf{shot} & \textbf{Methods} & \textbf{AG's News} & \textbf{Snippets} & \textbf{News Title}\\
\hline

\multirow{4}*{\tabincell{c}{5}} & BERT + FT & 32.30 & 35.30 & 31.31 \\

~ & PL & 73.20 & 72.01 & 66.54 \\

~ & KPT & 75.86 & \textbf{82.36} & 66.11 \\

~ & Ours & \textbf{78.53} & 82.03 & \textbf{71.88} \\

\hline

\multirow{4}*{\tabincell{c}{10}} & BERT + FT & 68.76 & 66.83 & 64.61 \\

~ & PL & 76.57 & 78.06 & 71.18 \\

~ & KPT & 78.54 & 81.68 & 70.56 \\

~ & Ours & \textbf{80.59} & \textbf{83.32} & \textbf{74.39} \\

\hline

\multirow{4}*{\tabincell{c}{20}} & BERT + FT & 76.63 & 77.65 & 71.54 \\

~ & PL & 80.45 & 82.05 & 74.33 \\

~ & KPT & 81.66 & 84.59 & 74.12 \\

~ & Ours & \textbf{83.35} & \textbf{86.33} & \textbf{75.87} \\

\bottomrule
\end{tabular}
\label{tab:booktabs}
\end{table}

\subsection{Parameter Sensitivity}

In this section, we investigate the influence of parameters in our proposed method, including top $N$ in concepts retrieval and the selected top $M$ words in verbalizer refinement. When we change one parameter, the rest others are fixed in the experiment. $N$ and $M$ are sampled from the set $\{$1,2,3,4,5,6,7$\}$ and $\{$30,40,50,60,70$\}$ respectively. All the results are reported in Figure 3 and Figure 4 respectively, and we set $N=5$ and $M=50$ to get the best and most stable results. It is worth mentioning that the results are declining when $M=80$, which indicates that more label words with weak relevance incur a performance loss, that is also why KPT achieved unsatisfying results in some datasets.

\begin{figure}[htbp]
\centering
\centerline{\includegraphics[width=3.5in,height=1.62in]{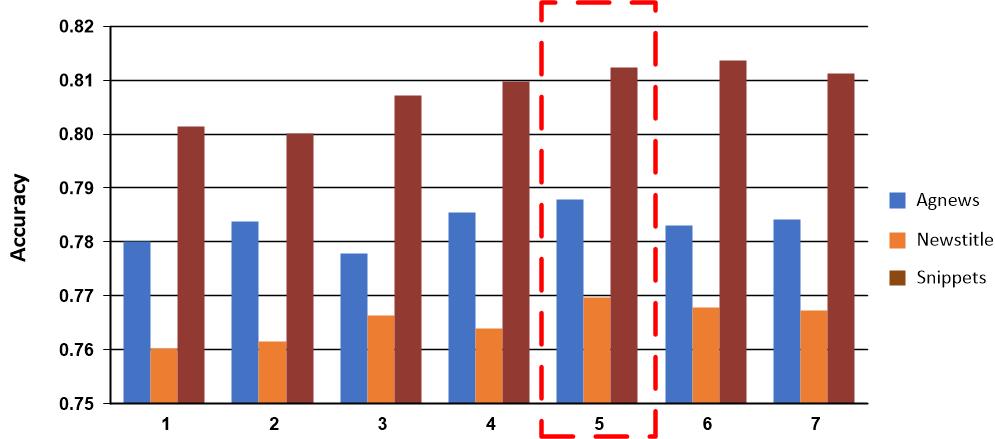}}
\caption{Parameter Influence of $N$ on all three datasets.}
\label{fig}
\end{figure}

\begin{figure}[htbp]
\centerline{\includegraphics[width=3.5in,height=1.62in]{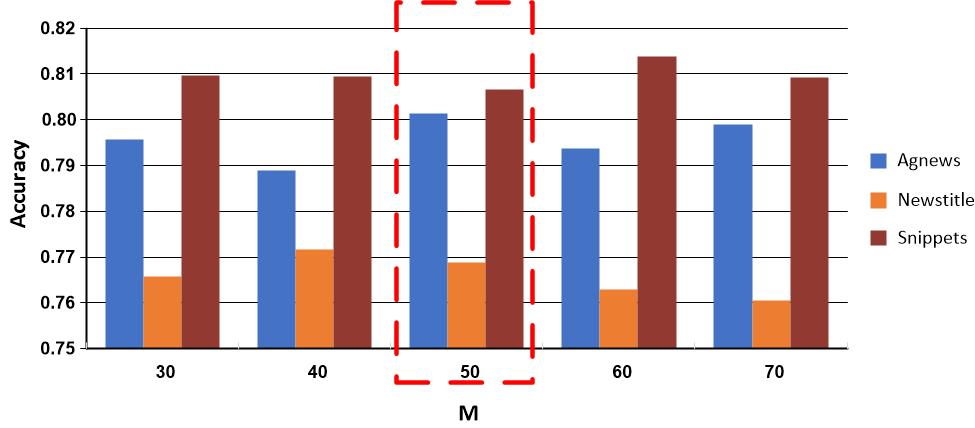}}
\caption{Parameter Influence of $M$ on all three datasets.}
\label{fig}
\end{figure}

%\begin{figure}[htbp]
%\centering
%\subfigure[]{
%\begin{minipage}[t]{0.5\linewidth}
%\centering
%\includegraphics[width=1.5in]{data_N.jpg}
%%\caption{fig1}
%\end{minipage}%
%}%
%\subfigure[]{
%\begin{minipage}[t]{0.5\linewidth}
%\centering
%\includegraphics[width=1.5in]{data_M.jpg}
%%\caption{fig2}
%\end{minipage}%
%}%
%\centering
%\caption{Parameter Influence of $N$ and $M$ on all three datasets.}
%\end{figure}

\subsection{Impact of Templates}

The templates have always been one of the important factors that affect the effectiveness of prompt-learning methods, we list all the templates used in our experiments in Table 4. The 10-shot experimental results of regular prompt-learning (PL), knowledgeable Prompt-tuning method (KPT), and our method with four templates on three datasets are reported in Table 5. The results reveal that our method has a significantly better performance on all three datasets than PL and KPT, our method can achieve substantial and consistent performance in all templates. In addition, we have observed that the experimental results of KPT are even worse than PL in the News Title and Snippets datasets, which shows that KPT cannot keep stable with the change of templates.

\begin{table}[htbp]
\begin{center}
\caption{The different templates on three datasets}
\begin{tabular}{ccc}
\hline
Dataset & id & Templates\\
\hline
\multirow{4}*{\tabincell{c}{AG news}} & 1 & A [mask] news : X\\

~  & 2 & X This topic is about [mask]\\

~  & 3 & The category of X is [mask]\\

~  & 4 & Topic:[mask] X\\

\hline

\multirow{4}*{\tabincell{c}{Snippets}} & 1 & X is about [mask]\\

~  & 2 & X This topic is about [mask]\\

~  & 3 & The category of X is [mask]\\

~  & 4 & The topic of X is [mask]\\

\hline

\multirow{4}*{\tabincell{c}{Newstitle}} & 1 & A [mask] news: X\\

~  & 2 & X This topic is about [mask]\\

~  & 3 & X is about [mask]\\

~  & 4 & The topic of X is [mask]\\

\hline

\end{tabular}
\end{center}
\end{table}

\begin{table}
\centering
\caption{The 10-shot results of accuracy with different templates on three datasets.}
\begin{tabular}{ccccc}
\toprule
\textbf{Methods} & \textbf{Template\_id} & \textbf{AG's News} & \textbf{Snippets} & \textbf{NewsTitle}\\
\hline

\multirow{5}*{\tabincell{c}{PL}} & 1 & 78.8 & 79.27 & 74.95\\

~ & 2 & 76.52 & 80.57 & 71.18 \\

~ & 3 & 72.99 & 75.64 & 69.1 \\

~ & 4 & 77.95 & 76.75 & 69.48 \\

%~ & avg & 76.57 & 78.06 & 71.18 \\

\hline

\multirow{5}*{\tabincell{c}{KPT}} & 1 & 80.29 & 81.68 & 72.6\\

~ & 2 & 77.0 & 82.13 & 70.98 \\

~ & 3 & 76.51 & 80.95 & 68.98 \\

~ & 4 & 78.54 & 80.99 & 69.68 \\

%~ & avg & 78.09 & 81.44 & 70.56 \\

\hline

\multirow{5}*{\tabincell{c}{ours}} & 1 & 79.76 & 83.32 & 76.59\\

~ & 2 & 77.68 & 82.55 & 73.82 \\

~ & 3 & 78.05 & 81.4 & 73.24 \\

~ & 4 & 80.59 & 81.98 & 73.92 \\

%~ & avg & 79.02 & 82.31 & 74.39 \\

\bottomrule
\end{tabular}
\label{tab:booktabs}
\end{table}

\subsection{Impact of Support Set}

In our experiments, 1000 instances of each category are sampled from the original training set to form a support set. In this section, to verify the impact of the support set, we altered the size from the set $\{$0,256,512,752,1000,1256,1512$\}$, the 10-shot experimental results with the first template on AG's News and Snippets dataset can be seen in Figure 5. From the observation of these results, our method keep the encouraging result on different sizes of support sets, which demonstrated the stability and effectiveness of our method.

\begin{figure}[htbp]
\centerline{\includegraphics[width=3in,height=1.67in]{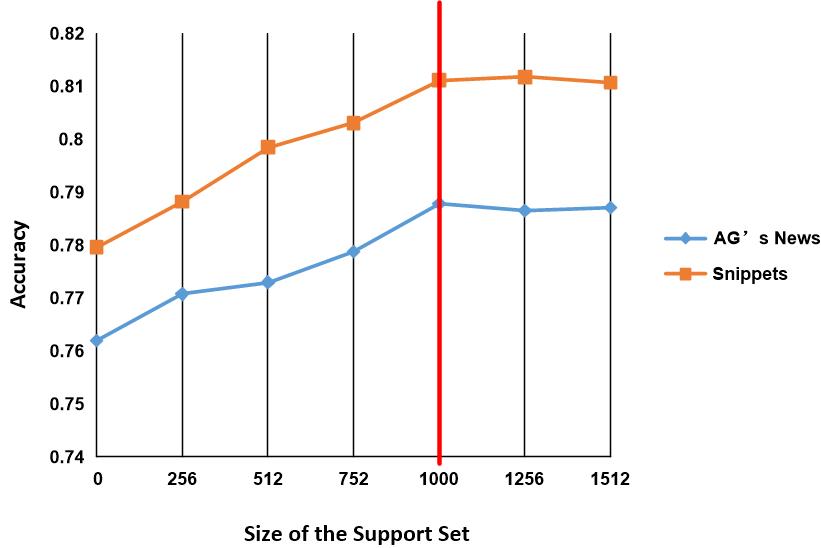}}
\caption{Impact of Support Set on AG's News and Snippets dataset.}
\label{fig}
\end{figure}

%\begin{table}
%\centering
%\begin{tabular}{ccc}
%\toprule
%\textbf{Support Size} & \textbf{AG's News} & \textbf{Snippets} \\
%\hline
%
%0 & 76.2 & 77.96\\
%
%256 & 77.08 & 78.82 \\
%
%512 & 77.3 & 79.85 \\
%
%752 & 77.88 & 80.31 \\
%
%1000 & \textbf{78.79} & 81.12 \\
%
%1256 & 78.65 & \textbf{81.18} \\
%
%1512 & 78.71 & 81.07 \\
%
%\bottomrule
%\end{tabular}
%\caption{Impact of Support Set on the 10-shot experiment with the first template in AG's News and Snippets datasets.}
%\label{tab:booktabs}
%\end{table}

\subsection{Effect of Our Method on Regular Text}

To evaluate the effect of our method on regular text, the supplement experiments are conducted on AG's News with all news titles and content. The average results are reported in Figure 6. Although our method is designed for short text classification on a few-shot scenario, our method achieved competitive results on regular text. The results show that the proposed method of concepts retrieval and refinement is more suitable for short text.

\begin{figure}[htbp]
\centerline{\includegraphics[width=3in,height=1.87in]{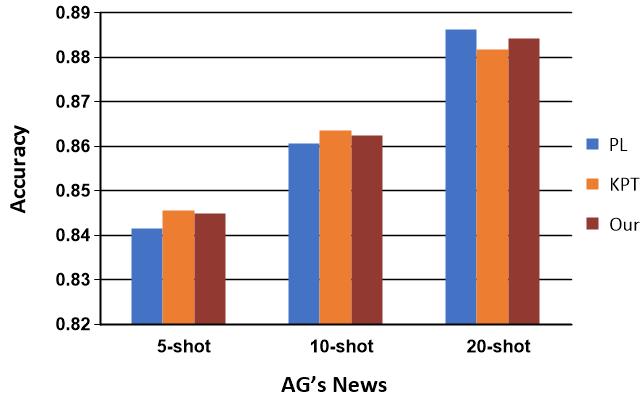}}
\caption{Effect of Our method on regular text in on AG's News Dataset with all news title and content.}
\label{fig}
\end{figure}

\section{Conclusion}

In this paper, we propose a prompt-learning strategy for short text classification. Taking the special characteristics of short text into consideration, the method can consider both the short text itself and class name during expanding label words space. The proposed method retrieves top $N$ concepts from the open Knowledge Graph and refines the expanded label words in embedding space. The experiments show the effectiveness of our method. We will extend our research work from the following two directions in the future. One is exploring better methods for automatic template construction and verbalizer designed on short text. The other is to incorporate more auxiliary information from external knowledge for some other tasks.

\section*{Acknowledgments}

This research is partially supported by the National Natural Science Foundation of China under grants (61906060,62076217), Yangzhou University Interdisciplinary Research Foundation for Animal Husbandry Discipline of Targeted Support (yzuxk202015), the Opening Foundation of Key Laboratory of Huizhou Architecture in Anhui Province under grant HPJZ-2020-02, Open Project Program of Joint International Research Laboratory of Agriculture and Agri-Product Safety (JILAR-KF202104).

\bibliographystyle{IEEEtran}
\bibliography{mybibfile}

\begin{thebibliography}{10}
\providecommand{\url}[1]{#1}
\csname url@samestyle\endcsname
\providecommand{\newblock}{\relax}
\providecommand{\bibinfo}[2]{#2}
\providecommand{\BIBentrySTDinterwordspacing}{\spaceskip=0pt\relax}
\providecommand{\BIBentryALTinterwordstretchfactor}{4}
\providecommand{\BIBentryALTinterwordspacing}{\spaceskip=\fontdimen2\font plus
\BIBentryALTinterwordstretchfactor\fontdimen3\font minus
  \fontdimen4\font\relax}
\providecommand{\BIBforeignlanguage}[2]{{%
\expandafter\ifx\csname l@#1\endcsname\relax
\typeout{** WARNING: IEEEtran.bst: No hyphenation pattern has been}%
\typeout{** loaded for the language `#1'. Using the pattern for}%
\typeout{** the default language instead.}%
\else
\language=\csname l@#1\endcsname
\fi
#2}}
\providecommand{\BIBdecl}{\relax}
\BIBdecl

\bibitem{sun2012short}
A.~Sun, ``Short text classification using very few words,'' in
  \emph{International ACM SIGIR conference on Research and development in
  information retrieval}, 2012, pp. 1145--1146.

\bibitem{cui2021pre}
Y.~Cui, W.~Che, T.~Liu, B.~Qin, and Z.~Yang, ``Pre-training with whole word
  masking for chinese bert,'' \emph{IEEE/ACM Transactions on Audio, Speech, and
  Language Processing}, vol.~29, pp. 3504--3514, 2021.

\bibitem{sriram2010short}
B.~Sriram, D.~Fuhry, E.~Demir, H.~Ferhatosmanoglu, and M.~Demirbas, ``Short
  text classification in twitter to improve information filtering,'' in
  \emph{International ACM SIGIR conference on Research and development in
  information retrieval}, 2010, pp. 841--842.

\bibitem{faqeeh2014cross}
M.~Faqeeh, N.~Abdulla, M.~Al-Ayyoub, Y.~Jararweh, and M.~Quwaider,
  ``Cross-lingual short-text document classification for facebook comments,''
  in \emph{International Conference on Future Internet of Things and Cloud},
  2014, pp. 573--578.

\bibitem{liu2010short}
Z.~Liu, W.~Yu, W.~Chen, S.~Wang, and F.~Wu, ``Short text feature selection for
  micro-blog mining,'' in \emph{International Conference on Computational
  Intelligence and Software Engineering}, 2010, pp. 1--4.

\bibitem{kim2014convolutional}
Y.~Kim, ``Convolutional neural networks for sentence classification. corr
  abs/1408.5882 (2014),'' \emph{arXiv preprint arXiv:1408.5882}, 2014.

\bibitem{lai2015recurrent}
S.~Lai, L.~Xu, K.~Liu, and J.~Zhao, ``Recurrent convolutional neural networks
  for text classification,'' in \emph{AAAI conference on artificial
  intelligence}, 2015.

\bibitem{chen2019deep}
J.~Chen, Y.~Hu, J.~Liu, Y.~Xiao, and H.~Jiang, ``Deep short text classification
  with knowledge powered attention,'' in \emph{AAAI Conference on Artificial
  Intelligence}, vol.~33, no.~01, 2019, pp. 6252--6259.

\bibitem{xu2019incorporating}
J.~Xu and Y.~Cai, ``Incorporating context-relevant knowledge into convolutional
  neural networks for short text classification,'' in \emph{AAAI Conference on
  Artificial Intelligence}, vol.~33, no.~01, 2019, pp. 10\,067--10\,068.

\bibitem{minaee2021deep}
S.~Minaee, N.~Kalchbrenner, E.~Cambria, N.~Nikzad, M.~Chenaghlu, and J.~Gao,
  ``Deep learning--based text classification: A comprehensive review,''
  \emph{ACM Computing Surveys (CSUR)}, vol.~54, no.~3, pp. 1--40, 2021.

\bibitem{yang2019data}
W.~Yang, Y.~Xie, L.~Tan, K.~Xiong, M.~Li, and J.~Lin, ``Data augmentation for
  bert fine-tuning in open-domain question answering,'' \emph{arXiv preprint
  arXiv:1904.06652}, 2019.

\bibitem{weng2020acquiring}
R.~Weng, H.~Yu, S.~Huang, S.~Cheng, and W.~Luo, ``Acquiring knowledge from
  pre-trained model to neural machine translation,'' in \emph{AAAI Conference
  on Artificial Intelligence}, vol.~34, no.~05, 2020, pp. 9266--9273.

\bibitem{qiang2020lexical}
J.~Qiang, Y.~Li, Y.~Zhu, Y.~Yuan, and X.~Wu, ``Lexical simplification with
  pretrained encoders,'' in \emph{AAAI Conference on Artificial Intelligence},
  vol.~34, no.~05, 2020, pp. 8649--8656.

\bibitem{goldberg2019assessing}
Y.~Goldberg, ``Assessing bert's syntactic abilities,'' \emph{arXiv preprint
  arXiv:1901.05287}, 2019.

\bibitem{ma2019universal}
X.~Ma, Z.~Wang, P.~Ng, R.~Nallapati, and B.~Xiang, ``Universal text
  representation from bert: An empirical study,'' \emph{arXiv preprint
  arXiv:1910.07973}, 2019.

\bibitem{jawahar2019does}
G.~Jawahar, B.~Sagot, and D.~Seddah, ``What does bert learn about the structure
  of language?'' in \emph{Annual Meeting of the Association for Computational
  Linguistics}, 2019, pp. 3651--3657.

\bibitem{han2021pre}
H.~Xu, Z.~Zhengyan, D.~Ning, G.~Yuxian, L.~Xiao, H.~Yuqi, Q.~Jiezhong,
  Z.~Liang, H.~Wentao, H.~Minlie \emph{et~al.}, ``Pre-trained models: Past,
  present and future,'' \emph{arXiv preprint arXiv:2106.07139}, 2021.

\bibitem{brown2020language}
T.~B. Brown, B.~Mann, N.~Ryder, M.~Subbiah, J.~Kaplan, P.~Dhariwal,
  A.~Neelakantan, P.~Shyam, G.~Sastry, A.~Askell \emph{et~al.}, ``Language
  models are few-shot learners,'' in \emph{Neural Information Processing
  Systems}, 2020, pp. 1877--1901.

\bibitem{gao2020making}
T.~Gao, A.~Fisch, and D.~Chen, ``Making pre-trained language models better
  few-shot learners,'' \emph{arXiv preprint arXiv:2012.15723}, 2020.

\bibitem{schick2020automatically}
T.~Schick, H.~Schmid, and H.~Sch{\"u}tze, ``Automatically identifying words
  that can serve as labels for few-shot text classification,'' \emph{arXiv
  preprint arXiv:2010.13641}, 2020.

\bibitem{schick2020exploiting}
T.~Schick and H.~Sch{\"u}tze, ``Exploiting cloze questions for few shot text
  classification and natural language inference,'' \emph{arXiv preprint
  arXiv:2001.07676}, 2020.

\bibitem{schick2020s}
------, ``It's not just size that matters: Small language models are also
  few-shot learners,'' \emph{arXiv preprint arXiv:2009.07118}, 2020.

\bibitem{hu2021knowledgeable}
S.~Hu, N.~Ding, H.~Wang, Z.~Liu, J.~Li, and M.~Sun, ``Knowledgeable
  prompt-tuning: Incorporating knowledge into prompt verbalizer for text
  classification,'' \emph{arXiv preprint arXiv:2108.02035}, 2021.

\bibitem{ding2021openprompt}
N.~Ding, S.~Hu, W.~Zhao, Y.~Chen, Z.~Liu, H.-T. Zheng, and M.~Sun,
  ``Openprompt: An open-source framework for prompt-learning,'' \emph{arXiv
  preprint arXiv:2111.01998}, 2021.

\bibitem{song2020sacpc}
C.~Song, X.-K. Wang, P.-f. Cheng, J.-q. Wang, and L.~Li, ``Sacpc: A framework
  based on probabilistic linguistic terms for short text sentiment analysis,''
  \emph{Knowledge-Based Systems}, vol. 194, p. 105572, 2020.

\bibitem{lu2013deep}
Z.~Lu and H.~Li, ``A deep architecture for matching short texts,''
  \emph{Advances in neural information processing systems}, vol.~26, pp.
  1367–--1375, 2013.

\bibitem{hu2009understanding}
J.~Hu, G.~Wang, F.~Lochovsky, J.-t. Sun, and Z.~Chen, ``Understanding user's
  query intent with wikipedia,'' in \emph{International Conference on World
  Wide Web}, 2009, pp. 471--480.

\bibitem{chen2020verbal}
J.~Chen, S.~Yan, and K.-C. Wong, ``Verbal aggression detection on twitter
  comments: Convolutional neural network for short-text sentiment analysis,''
  \emph{Neural Computing and Applications}, vol.~32, no.~15, pp.
  10\,809--10\,818, 2020.

\bibitem{wu2020mind}
F.~Wu, Y.~Qiao, J.-H. Chen, C.~Wu, T.~Qi, J.~Lian, D.~Liu, X.~Xie, J.~Gao,
  W.~Wu \emph{et~al.}, ``Mind: A large-scale dataset for news recommendation,''
  in \emph{Annual Meeting of the Association for Computational Linguistics},
  2020, pp. 3597--3606.

\bibitem{bollegala2018classinet}
D.~Bollegala, V.~Atanasov, T.~Maehara, and K.-i. Kawarabayashi,
  ``Classinet--predicting missing features for short-text classification,''
  \emph{ACM Transactions on Knowledge Discovery from Data (TKDD)}, vol.~12,
  no.~5, pp. 1--29, 2018.

\bibitem{hao2020chinese}
M.~Hao, B.~Xu, J.-Y. Liang, B.-W. Zhang, and X.-C. Yin, ``Chinese short text
  classification with mutual-attention convolutional neural networks,''
  \emph{ACM Transactions on Asian and Low-Resource Language Information
  Processing (TALLIP)}, vol.~19, no.~5, pp. 1--13, 2020.

\bibitem{yang2021hgat}
T.~Yang, L.~Hu, C.~Shi, H.~Ji, X.~Li, and L.~Nie, ``Hgat: Heterogeneous graph
  attention networks for semi-supervised short text classification,'' \emph{ACM
  Transactions on Information Systems (TOIS)}, vol.~39, no.~3, pp. 1--29, 2021.

\bibitem{adiwardana2020towards}
D.~Adiwardana, M.-T. Luong, D.~R. So, J.~Hall, N.~Fiedel, R.~Thoppilan,
  Z.~Yang, A.~Kulshreshtha, G.~Nemade, Y.~Lu \emph{et~al.}, ``Towards a
  human-like open-domain chatbot,'' \emph{arXiv preprint arXiv:2001.09977},
  2020.

\bibitem{ding2021prototypical}
N.~Ding, X.~Wang, Y.~Fu, G.~Xu, R.~Wang, P.~Xie, Y.~Shen, F.~Huang, H.-T.
  Zheng, and R.~Zhang, ``Prototypical representation learning for relation
  extraction,'' \emph{arXiv preprint arXiv:2103.11647}, 2021.

\bibitem{zhu2020incorporating}
J.~Zhu, Y.~Xia, L.~Wu, D.~He, T.~Qin, W.~Zhou, H.~Li, and T.-Y. Liu,
  ``Incorporating bert into neural machine translation,'' \emph{arXiv preprint
  arXiv:2002.06823}, 2020.

\bibitem{bao2020enhancing}
R.~Bao, J.~Wang, Z.~Zhang, and H.~Zhao, ``Enhancing pre-trained language model
  with lexical simplification,'' \emph{arXiv preprint arXiv:2012.15070}, 2020.

\bibitem{ye2020document}
Z.~Ye, G.~Jiang, Y.~Liu, Z.~Li, and J.~Yuan, ``Document and word
  representations generated by graph convolutional network and bert for short
  text classification,'' in \emph{European Conference on Artificial
  Intelligence}, 2020, pp. 2275--2281.

\bibitem{sun2019fine}
C.~Sun, X.~Qiu, Y.~Xu, and X.~Huang, ``How to fine-tune bert for text
  classification?'' in \emph{China national conference on Chinese computational
  linguistics}, 2019, pp. 194--206.

\bibitem{liu2021pre}
P.~Liu, W.~Yuan, J.~Fu, Z.~Jiang, H.~Hayashi, and G.~Neubig, ``Pre-train,
  prompt, and predict: A systematic survey of prompting methods in natural
  language processing,'' \emph{arXiv preprint arXiv:2107.13586}, 2021.

\bibitem{petroni2019language}
F.~Petroni, T.~Rockt{\"a}schel, P.~Lewis, A.~Bakhtin, Y.~Wu, A.~H. Miller, and
  S.~Riedel, ``Language models as knowledge bases?'' \emph{arXiv preprint
  arXiv:1909.01066}, 2019.

\bibitem{davison2019commonsense}
J.~Davison, J.~Feldman, and A.~M. Rush, ``Commonsense knowledge mining from
  pretrained models,'' in \emph{EMNLP-IJCNLP}, 2019, pp. 1173--1178.

\bibitem{han2021ptr}
X.~Han, W.~Zhao, N.~Ding, Z.~Liu, and M.~Sun, ``Ptr: Prompt tuning with rules
  for text classification,'' \emph{arXiv preprint arXiv:2105.11259}, 2021.

\bibitem{ding2021prompt}
N.~Ding, Y.~Chen, X.~Han, G.~Xu, P.~Xie, H.-T. Zheng, Z.~Liu, J.~Li, and H.-G.
  Kim, ``Prompt-learning for fine-grained entity typing,'' \emph{arXiv preprint
  arXiv:2108.10604}, 2021.

\bibitem{wang2019superglue}
A.~Wang, Y.~Pruksachatkun, N.~Nangia, A.~Singh, J.~Michael, F.~Hill, O.~Levy,
  and S.~R. Bowman, ``Superglue: A stickier benchmark for general-purpose
  language understanding systems,'' \emph{arXiv preprint arXiv:1905.00537},
  2019.

\bibitem{li2021prefix}
X.~L. Li and P.~Liang, ``Prefix-tuning: Optimizing continuous prompts for
  generation,'' \emph{arXiv preprint arXiv:2101.00190}, 2021.

\bibitem{lester2021power}
B.~Lester, R.~Al-Rfou, and N.~Constant, ``The power of scale for
  parameter-efficient prompt tuning,'' \emph{arXiv preprint arXiv:2104.08691},
  2021.

\bibitem{shin2020autoprompt}
T.~Shin, Y.~Razeghi, R.~L. Logan~IV, E.~Wallace, and S.~Singh, ``Autoprompt:
  Eliciting knowledge from language models with automatically generated
  prompts,'' \emph{arXiv preprint arXiv:2010.15980}, 2020.

\bibitem{hambardzumyan2021warp}
K.~Hambardzumyan, H.~Khachatrian, and J.~May, ``Warp: Word-level adversarial
  reprogramming,'' \emph{arXiv preprint arXiv:2101.00121}, 2021.

\bibitem{wei2022eliciting}
Y.~Wei, T.~Mo, Y.~Jiang, W.~Li, and W.~Zhao, ``Eliciting knowledge from
  pretrained language models for prototypical prompt verbalizer,'' \emph{arXiv
  preprint arXiv:2201.05411}, 2022.

\bibitem{zhang2015character}
X.~Zhang, J.~Zhao, and Y.~LeCun, ``Character-level convolutional networks for
  text classification,'' \emph{Advances in neural information processing
  systems}, vol.~28, pp. 649--657, 2015.

\bibitem{phan2008learning}
X.-H. Phan, L.-M. Nguyen, and S.~Horiguchi, ``Learning to classify short and
  sparse text \& web with hidden topics from large-scale data collections,'' in
  \emph{International Conference on World Wide Web}, 2008, pp. 91--100.

\bibitem{linmei2019heterogeneous}
H.~Linmei, T.~Yang, C.~Shi, H.~Ji, and X.~Li, ``Heterogeneous graph attention
  networks for semi-supervised short text classification,'' in
  \emph{EMNLP-IJCNLP}, 2019, pp. 4821--4830.

\bibitem{devlin2018bert}
J.~Devlin, M.-W. Chang, K.~Lee, and K.~Toutanova, ``Bert: Pre-training of deep
  bidirectional transformers for language understanding,'' \emph{arXiv preprint
  arXiv:1810.04805}, 2018.

\bibitem{liu2019roberta}
Y.~Liu, M.~Ott, N.~Goyal, J.~Du, M.~Joshi, D.~Chen, O.~Levy, M.~Lewis,
  L.~Zettlemoyer, and V.~Stoyanov, ``Roberta: A robustly optimized bert
  pretraining approach,'' \emph{arXiv preprint arXiv:1907.11692}, 2019.

\end{thebibliography}

\vfill

\end{document}